\documentclass[journal]{IEEEtran}

\usepackage{amsmath,amsfonts,amssymb,amsthm,mathtools}
\usepackage{color}
\usepackage{stmaryrd}
\usepackage{multirow}
\usepackage[numbers]{natbib}
\usepackage{fix-cm}
\usepackage{url}
\usepackage{times}
\usepackage{graphicx}
\usepackage{latexsym}
\usepackage[linesnumbered,ruled,vlined]{algorithm2e}
\usepackage{algpseudocode}
\usepackage{listings}
\usepackage{caption}
\usepackage{subcaption}
\usepackage{soul}
\usepackage{booktabs}

\newcommand{\ti}{I}
\newcommand{\inst}{\mathcal I}
\newcommand{\domn}{\mathcal D}
\newcommand{\grnd}{\mathcal G}

\theoremstyle{definition}
\newtheorem{defin}{Definition}

\definecolor{dkgreen}{rgb}{0,0.6,0}
\definecolor{gray}{rgb}{0.5,0.5,0.5}
\definecolor{mauve}{rgb}{0.58,0,0.82}

\newcommand{\stm}[3]{\left(\operatorname{#1},\operatorname{#2},\operatorname{#3}\right)}
\newcommand{\norm}[2]{\operatorname{#1}\left(#2\right)}
\newcommand{\name}[1]{$\operatorname{#1}$}
\newcommand{\namem}[1]{\operatorname{#1}}

\DeclareMathOperator*{\vals}{vals}

\makeatletter
\newcommand*{\rom}[1]{\expandafter\@slowromancap\romannumeral #1@}
\makeatother

\begin{document}

\title{Robby is Not a Robber (anymore): On the Use of Institutions for Learning Normative Behavior}

\author{%
  Stevan Tomic, %
  Federico Pecora
  and Alessandro Saffiotti%
\thanks{All authors are with the Center for Applied Autonomous Sensor
  Systems, School of Science and Technology, \"{O}rebro University,
  70187 \"{O}rebro, Sweden.  Contact author's e-mail:
  stevan.tomic@aass.oru.se;tomic01@gmail.com}% 
}

%\markboth{Journal of \LaTeX\ Class Files,~Vol.~14, No.~8, August~2015}%
%{Shell \MakeLowercase{\textit{et al.}}: Bare Demo of IEEEtran.cls for IEEE Journals}

\maketitle

\begin{abstract}
Future robots should follow human social norms in order to be useful and accepted in human society. In this paper, we leverage already existing social knowledge in human societies by capturing it in our framework through the notion of social norms. We show how norms can be used to guide a reinforcement learning agent towards achieving normative behavior and apply the same set of norms over different domains. Thus, we are able to: (1) provide a way to intuitively encode social knowledge (through norms); (2) guide learning towards normative behaviors (through an automatic norm reward system); and (3) achieve a transfer of learning by abstracting policies; Finally, (4) the method is not dependent on a particular RL algorithm. We show how our approach can be seen as a means to achieve abstract representation and learn procedural knowledge based on the declarative semantics of norms and discuss possible implications of this in some areas of cognitive science.
\end{abstract}

\begin{IEEEkeywords}
Norms, Institutions, Automatic Reward Shaping, Transfer of Learning, Abstract Policies, Abstraction, State-Space Selection, Schema
\end{IEEEkeywords}

%======================================================================
\section{Introduction}
%======================================================================
\label{sec.intro}

In order to be accepted in human society, robots need to comply with human social norms. The main goal of our research is to make robots capable of behaving in a socially-acceptable way, i.e., to adhere to social norms. In this paper, we do this by applying social norms in reinforcement learning (RL) settings. 

In recent years, reinforcement learning has achieved notable successes in simulated environments like classical ’Atari’ ~\citep{mnih2013playing} or board games e.g., go~\citep{silver2017mastering}, chess~\citep{silver1712mastering} and other computer games~\citep{vinyals2017starcraft},~\citep{OpenAIdota}. Additionally, significant successes have been achieved in real physical robotic control, after first learning policies in a simulated environment~\citep{andrychowicz2018learning}. Despite such impressive developments, RL is still mainly relegated to relatively tightly controlled settings. Using RL in real-world settings poses significant challenges, most notably, the credit assignment problem, which concerns how an action taken earlier influences the final reward, the size of state space, and the amount of training data. Also, the transfer of learned policies, to novel domains is not trivial to achieve. Challenges are complicated by the fact that robots have to act in the social space shared with humans. For example, in a learning process driven by exploration, an agent is not necessarily acting towards the achievement of a goal, which may result in possibly dangerous unexpected behaviors. Furthermore, learned behaviors and interaction with other agents or objects, may not be easily understandable or may seem odd to humans, thus making it difficult to realize human-level social interactions. 
Policies obtained via RL lead an agent to act in a way that maximizes the reward function, meaning that an agent may learn to achieve its goal in a way which is not socially acceptable. This somewhat Machiavellistic characteristic of RL agents indicates that policies that are to be used in social environments should be subject to, or at least biased by, social norms.

To make robotic RL agents behave socially, we aim to create policies that produce normative behavior depending on the social situation. In doing so, we also wish to avoid hand-crafting policies (or the decision-theoretic planning~\citep{Boutilier:1999:DPS:3013545.3013546} models), that is, provide agents with principles that can be applied across domains and boundary conditions.
We combine specification and learning of social norms, so that these social norms can be applied in a general and reusable way to agents that learn, resulting in learned policies that generate normative behaviors.
Towards this end, we propose the use of {\em institutions} (e.g. ~\citep{searle2005institution},~\citep{north1990institutions},~\citep{ostrom2009understanding}). Institutions give a social dimension to the actions of agents and provide regulative mechanisms to guide agents in their interactions towards normative behaviors. The advantage of using institutions is twofold: (A) They provide {\em social knowledge} by capturing the set of norms that should be followed given the current situation; (B) Norms in the institution are usually {\em abstracted} from the domain in which they can be applied, thus the same set of norms can be applied in different domains.

In this paper, we leverage the fact that our institutional model can distinguish between socially acceptable and not acceptable behaviors by formally verifying them against norm semantics and by exploiting the facts (A) and (B). As we will see, (A) can be used to automatically create a reward system that can guide RL towards policies that adhere to norms, addressing the well-known credit assignment problem, while (B) will allow us to create abstracted normative policies, addressing state-space reduction with automatic state-space/feature selection, which can be applied in different domains without re-learning. Finally, the framework is
agnostic to the particular learning / RL approach.

We start with a brief look at the related work, followed by a description of the basics of our institutional framework. Then we cover the background of RL approaches and describe how we used our framework in RL settings. A set of experiments are then presented as proofs of concept that: (1) normative policies can indeed be learned; (2) an automatic reward system based on norms can be created that helps robotic agents learn in the right direction; (3) policies can be abstracted and applied to different domains; and (4) our approach can be used in multi-agent settings. Finally, we discuss our achievements with the focus on abstract knowledge representation and its possible further implications. 

%======================================================================
% EOF
%======================================================================

%======================================================================
\section{Related Work}
%======================================================================
\label{sec.related}

\textit{Classical Problems in RL.} One of the well-known problems in RL, the {\em the credit assignment problem}~\citep{minsky1963computers}, decreases learning ability for long temporal horizons with sparse rewards. {\em The curse of dimensionality}~\citep{bellman1957dynamic} is another problem where the exploration of states has to increase exponentially with the increase of state space. Social robots face these problems since they should act in a physical world with a complex state-space and long temporal horizons. Finally, the transfer of learning -- how learned knowledge can be reused for novel domains -- is notoriously difficult.
Several ways to address these problems have been reported in the literature. RL agents can receive intermediate rewards which serve as `progress estimators' (indicators), which reduces exploration and directs an agent's behavior towards the goal. ~\citet{mataric1997reinforcement} shows how this can significantly reduce learning time. \citet{ng1999policy} analyzed `reward shaping' mechanisms and a potential-shaping function that can speed up convergence without changing the previous optimal policy. The major shortcoming of this approach is that such intermediate feedback signals require significant engineering effort.
Hierarchical RL (HRL) provides a different approach to address long temporal horizons. In such methods, the domain (or temporal aspect of it) are represented at another level of abstraction~\citep{sutton1999between}. HRL methods mitigate the temporal assignment problem: by shortening the temporal horizon, better (structural) exploration is achieved, while sub-policies enable transfer learning.
%achieve better (structural) exploration by using sub-policies and provide the possibility for transfer of learning.
%, where sub-policies can be trained for a novel domain/problem, leaving the higher structures intact.  
A popular way to address the `curse of dimensionality' problem, especially in robotics, is to reduce a state space to a smaller one. Behavior-based systems (BBS)~\citet{mataric2001learning} encapsulate behaviors as higher-level structures combining sensing and action, which helps in reducing the underlying state-space. Robots are then trained over such higher-level state space. The state-space can be reduced, for example, by defining the measure of importance of a sensor~\citep{kishima2013decision} or by learning connections between large perceptual states spaces and hidden states~\citep{mccallum1996learning}. Similarly, the number of agent actions can also be reduced, where the most popular approach is to do that with the `option' framework of~\citet{sutton1999between}. Options can be understood as `closed-loop' sub-policies with certain duration depending on the terminating conditions and can combine more primitive agent actions. In general, state-abstraction and action abstraction are usually studied separately. 

Both reward shaping and state-space/action reduction are crucial ideas that are utilized in our work. The way we define the semantics of norms and the way they are associated with social structures, allows us to naturally abstract both the state-space and the action space of learning agents. Such abstraction reduces the dimensionality problem while at the same time addresses the transfer learning problem so that we can apply the same norms over novel domains. Furthermore, by utilizing the process of formal verification, we can assign normative states to agent behaviors such as fulfillment or violation. Consequently, behavior can then be evaluated during execution to define a normative shaping function and to mitigate the credit assignment problem. 

%\snote{Related Work - NORMS}
\textit{Norms.} In the context of RL, learning human normative behavior is usually achieved through inverse reinforcement learning (IRL) algirithms~\citep{ng2000algorithms}. 
%Similar approaches are also used in robotic (Inverse optimal control) ~\citep{finn2016guided}.  
The motivation behind this area of research is to learn (infer) a reward distribution from the exemplary behavior of expert humans. The obtained reward distribution is used in ordinary RL settings to train artificial agents, where the resulting policies lead to the desired behavior. While this approach does not learn norms directly, an additional proposal explicitly represents norms which are then used to infer them from demonstrated behaviors~\citep{kasenberg2018norms}. 
% Another research argues how non-Markov rewards (rewards depending on history) can be translated to non-Markovian tasks can be translated.
Additionally, agents can also be trained to learn behavior based on human feedback~\cite{christiano2017deep}. 

In our approach, we leverage given social knowledge captured in the form of social institutions. Our framework enables us to formally represent this knowledge and use it for learning normative behavior in RL settings, addressing some of the classical RL problems described above.

\textit{Institutions.} The concept of institution stems from economic and social sciences~\citep{searle2005institution}. The concept is loosely defined across different fields. According to \citet{north1990institutions} institutions are, simply defined as {\em ``the rules of the game in a society''}, while~\citet{ostrom2009understanding} defines them as: {\em ``the prescriptions that humans use to organize all forms of repetitive and structured interaction...''}. In multi-agent systems (MAS) institutions are usually seen as a set of norms. The institution and its norms are used to coordinate organizations of agents, thus they are sometimes called ``coordination artifacts''~\citep{Silva2008a}. 
%Our notion of institutions capture prototypical social knowledge, expressed thorough norms, providing an idealized view on how agents should behave. 
One of the advantages of using institutions to organize agents is that they are typically abstracted from the particular agent domain. Thus, ideally, the same institution should be able to organize different heterogeneous agents in various environments. 

In our work, we use the concept of institution to formalize the normative aspects of a social situation (context) by encapsulating a set of abstract norms. The core of learning normative behaviors is concerned with the particular relations of state values in an agent's execution trace (norms semantic functions), which {\em `count as'}~\citep{searle2005institution} institutional norms. The count-as principle is explored in MAS~\citep{aldewereld2010making}. We employ a mapping between institutional concepts and a domain, called {\em grounding}. This mapping, realizes the count-as concept, lifting an agent's ``raw'' execution traces to their institutional (social) meaning and binding them to the normative dimension through obligations, permissions or prohibitions.

%======================================================================
% EOF
%======================================================================

%======================================================================
\section{The Formal Model of Institution}
%======================================================================
\label{institutions.sec}

We start by introducing our formal institution framework. The framework is presented in full detail
in~\citep{tomic2018norms}, while in this section we only present the parts of that
framework that are relevant to this paper.

%----------------------------------------------------------------------
\subsection{Institutions}
%----------------------------------------------------------------------
\label{institutions.sub}

Institutions encapsulate a collection of \emph{norms} together with the \emph{roles},  \emph{actions} and \emph{artifacts} that these norms
refer to.  For example, the desired behaviors of agents participating in a store (marketplace) can be modeled, where its norms include relations between sellers, customers, goods. A norm in this institution could state, for instance, that a buyer should pay for certain goods to a certain seller. We define the following sets:

{\small
\begin{align*}
	\textit{Roles} &= \{role_1, role_2, \dots , role_m\} \\ 
	\textit{Arts} &= \{art_1, art_2, \dots , art_a\} \\ 
	\textit{Acts} &= \{act_1, act_2, \dots , act_k\}
\end{align*}
}
Examples of \textit{Roles} are a `customer' in a store or a `goal-keeper' in a football game.  \textit{Arts} are the artifacts of the institution, e.g., `goods' or a `cash register'.  \textit{Acts} are actions that can be performed, e.g., `exit' a location or `pay' for a carton of milk. We define norms to be predications over statements:

\begin{defin} A {\em norm} is a statement of the form $q(trp^*)$, where
  $q$ is a qualifier and $trp$ is a triple of the form:
  \begin{align*}
    trp \in \textit{Roles} \times \textit{Acts} \times (\textit{Arts} \cup \textit{Roles})
  \end{align*}
\end{defin}
\noindent 
Qualifiers can be unary relations like \name{must} or \name{must-not}, or n-ary ones like \name{inside} or \name{before}.  For example, \name{must} can be used to represent the obligation `A buyer \textit{must} pay with cash': 
$$\norm{must}{\stm{Buyer}{Pays}{Cash}}.$$ 
A qualifier \name{in\_front\_of} can specify a spatial constraint on the location of an action, as in `Paying should be performed \emph{in front of} a cash register': 
$$\norm{in\_front\_ of}{\stm{Buyer}{Pays}{CashRegister}}.$$ 
Binary qualifiers can express relations between statements, e.g., temporal relations such as \name{before} or \name{during}, for instance 
$$\norm{before}{\stm{Buyer}{GetGoods}{Goods},\stm{Buyer}{Pays}{Cash}}$$
indicating that `A buyer should get the goods \emph{before} paying'.
% For clarity, for all qualifiers representing necessity, we denote a
% subset of norms called \emph{obligation norms}.

Institutions put all the above elements together:

\begin{defin} An {\em institution} is a tuple
\begin{align*}
\inst = \langle \textit{Arts}, \textit{Roles}, \textit{Acts}, \textit{Norms} \rangle.
\end{align*}
% where
% $\textit{Norms} = q(trp^*)$.
\end{defin}

\noindent
This definition embodies a pragmatic view of institutions, which is in accordance to North's view that institutions constitute ``the human-devised constraints that shape social interaction''~\cite{north1990institutions}.

%----------------------------------------------------------------------
\subsection{Domains}
%----------------------------------------------------------------------
\label{domain.sub}

An institution is an abstraction, which can be instantiated in concrete systems that are physically different but are described by the same structure.  For instance, the same `store' institution can be used to describe or regulate behaviors of agents in different markets, irrespective of these agents being humans, robots, or a combination of both.  We call such a concrete system a \emph{domain}.

The domain ($\domn$) is characterized by: the set of agents $A$ that can include humans (e.g., \name{john}), robots (e.g., \name{robby}), or both; the collection $B$ of all behaviors that these humans or robots can perform, like \name{pick} or \name{speak};  a set $O$ of objects in the domain, like \name{door}, \name{battery}. In addition, a domain includes a set $R = \{\rho_1 \dots , \rho_n \}$ of state variables. Each state variable $\rho_i$ defines a property pertaining to an entity (agent, behavior, or object) in the domain. For instance, $\namem{pos}(\namem{robby})$ is a state variable that indicates the position of the agent \name{robby}, while
$\namem{active}(\namem{pick},\namem{robby})$
indicates the activation of \name{robby}'s behavior \name{pick}.
We denote with $\vals(\rho)$ the set of possible values of state variable $\rho$, e.g., 
$\vals(\namem{active}(\namem{pick},\namem{robby})) = \{\top, \bot \}$.

\begin{defin}
\label{def:stspace}
  The \emph{state space} of $\domn$ is $\mathcal{S} = \prod_{\rho \in R} \vals(\rho)$,
  where $\vals(\rho)$ are the possible values of state-variable $\rho$.
  We call any element $s \in \mathcal{S}$ a \emph{state}.  The value of
  $\rho$ in state $s$ is denoted $\rho(s)$.
\end{defin}
\noindent 
In a dynamic environment, the values of most properties change over time.
In our formalization, we represent time points by natural numbers in
$\mathbb{N}$, and time intervals by pairs $\ti = [t_1, t_2]$ such that
$t_1, t_2 \in \mathbb{N}$ and $t_1 \leq t_2$.  We denote by $\mathbb{I}$
the set of all such time intervals.

\begin{defin}
\label{def:traje}
A \emph{trajectory} is a pair $(\ti,\tau)$, where $\ti \in \mathbb{I}$
is a time interval and $\tau : \ti \to \mathcal{S}$ maps time to states.
\end{defin}
\noindent In simple words, a trajectory represents how the domain evolves over a
given time interval.

%----------------------------------------------------------------------
\subsection{Grounding}
%----------------------------------------------------------------------

\begin{defin}
  Given an institution $\inst$ and a domain $\mathcal{D}$, a
  \emph{grounding} of $\inst$ into $\mathcal{D}$ is a tuple
  $\mathcal{G} = \langle \mathcal{G}_A, \mathcal{G}_B, \mathcal{G}_O
  \rangle$, where:
%  {\small
  \begin{itemize}
  \item ${\mathcal G}_A \subseteq Roles \times A$ is a \emph{role grounding},
  \item ${\mathcal G}_B \subseteq Acts \times B$ is an \emph{action grounding},
  \item ${\mathcal G}_O \subseteq Arts \times O$ is an \emph{artifact grounding}.
  \end{itemize}
%  }
\end{defin}
\noindent 
Grounding plays an important role in our framework, by establishing the relation between an abstract institution and a specific domain.  Grounding provides the key to reuse the same abstract institution to describe or regulate different systems.  For example, the 'store' institution can be grounded in analogous environments, e.g. different physical marketplaces, using different ${\mathcal G}$'s. Also, different institutions can be grounded to the same domain. 

Grounding directly depends on object affordances. For example, if a grounding maps an agent $a$ to a given role, and this role involves the execution of a certain action, then $a$ must be able to perform a behavior that grounds that action. For more details please refer to ~\citep{tomic2018norms}. In this paper we assume that all groundings are admissible. 

%----------------------------------------------------------------------
\subsection{Norms Semantics and States}
%----------------------------------------------------------------------
\label{sub.semantics}

By grounding an institution $\inst$ into a domain $\domn$, we impose
that the norms stated for abstract entities in $\inst$ must hold for the
corresponding concrete entities in $\mathcal{D}$.  But what does ``hold''
mean here?  To answer this question, we give norms semantics in terms of
executions in a physical domain: intuitively, the semantics of norm is
the set of all trajectories that comply with the condition that the norm
is meant to express.  For example, the semantics of the norm
$\norm{must}{\stm{buyer}{getGoods}{goods}}$ is given by all
trajectories where the behavior $b$, to which the action \textit{getGoods}
is grounded, is executed (at least once) by any agent $a$, to which the role
\textit{buyer} is grounded. 

Formally, let $T$ be the set of all possible trajectories $(\ti,\tau)$
over the state variables of domain $\mathcal{D}$.  Given a norm
$q(trp^*)$, we define its semantic $\llbracket q(trp^*) \rrbracket$ in
$\mathcal{D}$ as:
\begin{align*}
\llbracket q(trp^*) \rrbracket \subseteq T
\end{align*}

\noindent
We differentiate between the following type of norm semantics: 
Fulfillment semantics, $\llbracket q(trp^*) \rrbracket_F$, that defines the set of trajectories fulfilling the norm; and violation semantics, $\llbracket q(trp^*) \rrbracket_V$, expressing the set of trajectories violating the norm. Naturally a trajectory cannot be an element of both fulfillment and violation semantics for a given norm, that is: $\llbracket q(trp^*) \rrbracket_F \cap \llbracket q(trp^*) \rrbracket_V = \emptyset$.
%
%and not-fulfilled semantics, $\llbracket q(trp^*) \rrbracket_{NF}$, as a set of all other trajectories: $\{ \llbracket q(trp^*) \rrbracket_{NF} \} = \{T\} \setminus \{ \llbracket q(trp^*) \rrbracket_{F} \} \setminus \{\llbracket q(trp^*) \rrbracket_{V} \}$
%
Accordingly, we define a function that indicates the {\em state of a norm} given a trajectory as follows:
$ns( q(trp*), (\ti,\tau) ) = \{f, v, n\}$, 
where $f$ stands for {\em fulfilled}, $v$ for {\em violated} and $n$ for {\em neutral} state, neither fulfilled nor violated:

%{\small
%\vspace{-5mm}
\begin{equation*}
ns := 
\begin{cases}
f, 	\text{iff } (\ti,\tau) \in \llbracket q(trp^*) \rrbracket_{F}\\
v, 	\text{iff } (\ti,\tau) \in \llbracket q(trp^*) \rrbracket_{V}\\
n, \text{otherwise}
\end{cases}
\end{equation*}
%}

\paragraph{Examples of Semantics.}\label{sem.para} A variety of norm semantics can be expressed as constraints on trajectories, i.e.,
as constraints on the possible values of state variables over time.
% 
% For example, the fact that some behavior should be executed only in
% certain places can be expressed by binding the state variable
% \name{active} of that behavior to the \textit{pos} of the place in
% question.
% 
For example, the fulfillment semantics of the \name{must} norm can be expressed by requiring that the
state variable \name{active} of the relevant behavior is true at least
once.  Formally:
%%% MUST
%
\begin{equation*}
\label{sem.must}
\begin{split}
   \llbracket &\norm{must}{(role,act,art)} \rrbracket_F \equiv\\
   \{ &(\ti,\tau) \mid \forall a \in A_{role} . \exists (b,t) \in B_{act} \times \ti : \\
   &\namem{active}(b,a)(\tau(t)) = \top \},
\end{split}
\end{equation*}
where $A_{role}$ is the set of all agents $role$ is grounded to, and $B_{act}$ is the set of all behaviors $act$ is grounded to.  Intuitively, these semantics select all trajectories where any agent taking $role$ activates at least once a behavior that performs $act$. Variants of this semantics are possible, for example, stating that the behavior should be enacted at all times (not just once in the trajectory). In more demanding scenarios, a state variable indicating 'successful execution' can be used instead of simpler 'activation'. Notice that $art$ in the norm is ignored, since it is not of interest for the semantic definition of this norm. However, it may be relevant in other norms, as in the following example:
%%% USE
%
\begin{equation*}
\label{sem.use}
\begin{split}
\llbracket &\norm{use}{(role,act,art)} \rrbracket_F \equiv\\
\{ &(\ti,\tau) \mid \forall (b,a,t) \in B_{act} \times A_{role} \times \ti . \exists o \in O_{art} : \\
&(\namem{active}(b,a)(\tau(t)) = \top \implies \namem{usedObj}(b,a) = o)  \},
\end{split}
\end{equation*}

\noindent
where $O_{art}$ is the set of all objects. The semantics states that all agents in the role grounding with behaviors in the action grounding that are active should use the same object grounded to artifacts. Sometimes it is useful to join the semantics of the norm \name{must} and the norm \name{use}:
%%% MUSTUSE
%
\begin{equation*}
\label{sem.mustuse}
\begin{split}
\llbracket &\norm{mustUse}{(role,act,art)} \rrbracket_F \equiv\\
\{ &(\ti,\tau) \mid \forall a \in A_{role} . \exists (b,o,t) \in B_{act} \times O_{art} \times \ti : \\
&\namem{active}(b,a)(\tau(t)) = \top \wedge (\namem{active}(b,a)(\tau(t)) = \top \implies \\
& \namem{usedObj}(b,a)(\tau(t)) = o))  \},
\end{split}
\end{equation*}

\noindent
The semantics selects all trajectories in which grounded agents with corresponding grounded behaviors are executed, and such execution implies that grounded objects are used. Similarly spatial and temporal semantics can be defined. The next example shows obligatory $at$ semantics:
%%% MUSTAT
%
\begin{equation*}
\label{sem.mustat}
\begin{split}
\llbracket &\norm{mustAt}{(role,act,art)} \rrbracket_F \equiv\\
\{ &(\ti,\tau) \mid \forall a \in A_{role} . \exists (b,o,t) \in B_{act} \times O_{art} \times \ti : \\
&\namem{active}(b,a)(\tau(t)) = \top \wedge (\namem{active}(b,a)(\tau(t)) = \top \implies \\
& \namem{position}(b,a)(\tau(t)) = \namem{position}(o)(\tau(t)))  \},
\end{split}
\end{equation*}

\noindent  
An example of temporal semantics is the following:

\begin{equation*}
\label{norm.before}
\begin{split}
\llbracket &\norm{before}{(role_1,act_1,art_1),(role_2,act_2,art_2)} \rrbracket_F \equiv\\
\{ &(\ti,\tau) \mid \forall (a_1, a_2) \in A_{role_1} \times A_{role_2}, \\
&(b_1, b_2) \times B_{act_1} \times B_{act_2}, \forall (t_1, t_2) \in \ti \times \ti,  \\
&(\namem{active}(b_1,a_1)(\tau(t_1)) = \top \wedge \namem{active}(b_2,a_2)(\tau(t_2)) = \top\\
&\implies t_1 < t_2) \wedge \\
& (\forall (a_1, a_2) \in A_{role_1} \times A_{role_2}, (b_1, b_2) \times B_{act_1} \times B_{act_2}, \\
& \exists (t_1, t_2) \in \ti \times \ti: \\
& \namem{active}(b_1,a_1)(\tau(t_1)) = \top \wedge \namem{active}(b_2,a_2)(\tau(t_2)) = \top) \}. \\
\end{split}
\end{equation*}
Simply said, `\name{before}' fulfill semantics declares that all grounded agents behaviors for all grounded agents, have to be in a certain order: the first should precede the second, and both have to be active at a certain time point in a trajectory. While for spatial semantics the states $n$ and violated $v$ are practically the same in most cases, for temporal semantics it is useful to explicitly define violation semantics, since such norms can often be in $n$ state thus neither violated nor fulfilled. One simple way to define violation semantics for the \name{before} norm, $\llbracket \norm{before}{(\cdots)} \rrbracket_V $, would be to switch the non-equality symbol between 
time points $t_1$ and $t_2$ to $t_1 \geq t_2$, and to ensure the activation of the 'second' behavior:

\begin{equation*}
\label{norm.before}
\begin{split}
\llbracket &\norm{before}{(role_1,act_1,art_1),(role_2,act_2,art_2)} \rrbracket_V \equiv\\
\{ &(\ti,\tau) \mid \forall (a_1, a_2, b_1, b_2) \in A_{role_1} \times A_{role_2} \times B_{act_1} \times B_{act_2} \\
&\forall (t_1, t_2) \in \ti \times \ti  \\
&(\namem{active}(b_1,a_1)(\tau(t_1)) = \top \wedge \namem{active}(b_2,a_2)(\tau(t_2)) = \top\\
&\implies t_1 >= t_2)  \wedge \exists t_{2}' \in \ti:\\
&\namem{active}(b_2,a_2)(\tau(t_2')) = \top \wedge t_2=t_2' \}
\end{split}
\end{equation*}

\noindent
Note that the semantics does not ensure existence of $(\namem{active}(b_2,a_2)(\tau(t_1'))$, since if $b_2$ is active at $t_2$ but $b_1$ is not active at any $t_1 < t_2$, the norm is already violated. 
When a trajectory is neither fulfilled nor violation, the trajectory is in a neutral state ($n$). As we will see, this helps us shape the reward function and provide informative feedback immediately when the norm state becomes either fulfilled or violated.

Other types of norms and their semantics can be defined depending on
the application. The expressiveness of norms depends on two factors, namely on the complexity of relations in norms semantics and on the availability of appropriate state variables $R$. It is important to stress that semantics is given in the terms of execution and is abstracted, since it is defined through grounding to abstract institutional concepts.

An important property of our semantics, which is exploited in this paper, is that semantics is defined over state-space variables belonging in a sub-sets of equivalence classes in the state-space. For example, semantics of the norm `\name{must}' is defined over elements in the equivalence class of the state variables indicating activation of an agent's behavior: 
$$[active] = \{active(a,b) \in R \mid a \in A, b \in B \}$$
Then, a particular state-variable, used in norm semantics, is decided by grounding. 
%In this way we can classify our state-variables over particular feature dimensions based on such equivalence classes. 

%----------------------------------------------------------------------
\subsection{Adherence}
%----------------------------------------------------------------------
\label{sub.adherence}

We are now in a position to define what it means for a given physical
system to behave in compliance with an institution.  Consider an
institution $\inst$, and suppose that $\inst$ has been grounded on a
given domain $\domn$ through some grounding $\mathcal{G}$.
Further suppose that all the norms in $\inst$ are given fulfillment semantics in
$\mathcal{D}$ through the function $\llbracket \cdot \rrbracket_F$.

The following definition tells us whether or not a specific, concrete
execution in $\domn$ \emph{adheres to} the abstract institution
$\inst$ given the above grounding and semantics.

\begin{defin}
\label{def:adheres}
  A trajectory $(\ti,\tau)$ {\em adheres} to an institution $\inst$, with
  admissible grounding $\mathcal{G}$ and semantics function $\llbracket
  \cdot \rrbracket_F$, if 
\begin{align*}
  (\ti,\tau) \in \llbracket \namem{norm} \rrbracket_F, \forall \namem{norm} \in \textit{Norms}.
\end{align*}
\end{defin}
With the adherence property, we can distinguish between trajectories (executions) which are normative (socially acceptable) and others which are not. To do that, we rely on the formal verification mechanisms described in our previous work~\citep{tomic2018norms}. With verification, we can evaluate a trajectory and say if it is adherent (to an institution) or not. Moreover, we can also evaluate the status of each norm separately during agent execution. Such states are then used to automatically create a reward function.

%----------------------------------------------------------------------
\subsection{From Declarative to Procedural Knowledge}
\label{sub.ontology}
%----------------------------------------------------------------------
The ontology in our framework is represented with several layers.
The institutional layer specifies abstract norms, through relations between social-level categories (roles, actions, artifacts) providing information about `what' ought to be normative behavior. 
By associating an institution (abstract layer) with the domain (concrete) layer, we give social meaning to agents behaviors and interactions through declarative norm semantics in terms of execution. This association is realized with the middle layer with the notion of grounding. It binds institution-level norms with the concrete domain elements (agents, behaviors, objects). 

The next section considers answering the question `how': In which way and when exactly should agents act so that their behavior is socially acceptable in the context of a given institution and grounding. While in our previous research we have achieved synthesizing adherent trajectories with model-based planners~\cite{wasik2018towards},~\cite{tomic2018norms},~\cite{tomic2018towards}, in this paper, we focus our attention on answering this question with reinforcement learning algorithms. 

%======================================================================
% EOF
%======================================================================

%======================================================================
\section{Applying Institution Models to RL agents}
%======================================================================
\label{sec.rl}

%\snote{general intro to RL, mdps, states, values, action values,  rewards, function approximation...}
Markov Decision Processes (MDPs) are an example of how reinforcement learning can be used to direct the actions of agents. A MDP is defined as a tuple $<\mathcal{S},\mathcal{A},\mathcal{P},\mathcal{R},\gamma>$, where $\mathcal{S}$ is the state space, $\mathcal{A}$ is the finite set of actions, $\mathcal{P}$ is the model given as transitions probability between states depending on actions, and $\mathcal{R}$ is a set of rewards \--- a scalar feedback signal which agents get as they change their states. $\gamma \in (0,1)$ is a discount factor indicating the importance of future rewards. The goal of a RL agent (the RL problem) is to find a mapping between states and actions which maximizes the amount of reward the agent receives, known as cumulative reward or return. Such a mapping represents a policy $\pi(a_t \mid s_t)$, which gives the probability of taking action $a_t \in \mathcal{A}$ given the state $s_t \in \mathcal{S}$. In deterministic policies, this probability is equal to $1$.
%Policy can be stochastic or deterministic...
%In MDP policies depend on the current state of the agent and not the history, due to its Markov property.
RL algorithms use the notion of value functions: a value function $V^{\pi}(s)$ provides the expected return of rewards starting from state $s$ acting according to policy $\pi$; and/or a action-value function (Q-function), $Q^{\pi}(s,a)$, which estimates the value of a state given the action. Knowing the optimal $Q$ function ($Q^*$) gives us an optimal policy for an agent for any state $s$, that is, $\underset{a \in \mathcal{A}}{\mathrm{argmax}} \; {Q^*(s,a)} $. RL algorithms update the value function either from sampled data, if the model $\mathcal{P}$ is not-known (model-free RL), or by leveraging on the provided model, where it is possible to use, e.g., dynamic programming (model-based RL). Value updates propagate the truth about the rewards until they converge to optimal values. For example, a standard idea in RL to learn Q-values is known as temporal difference learning:

$Q(s,a) \leftarrow Q(s,a) + \alpha (r + \gamma  \; \underset{a'}{\max} \; Q(s',a') - Q(s,a))$, 

\noindent
where $\alpha$ is a learning rate, and $s'$ and $a'$ are the state and action subsequent to $s$ and $a$. For a variety of real world problems the number of states may be too large to be stored (a tabular representation), and calculating values for each state may not be feasible. Value function approximation methods addresses this problem. The idea is to generalize over (semantically) similar states across a distributed representation of a state-space with fewer parameters. 
In such an approach, state-space variables are represented as a feature vector, where each feature is a number, describing some property in the state-space, e.g., the position of an agent along x coordinate, the activation of an agent's behavior, etc.
%$\teta$, such that $\hat{Q}(s,\teta) \approx {Q}(s)$. 
The policy ($\pi_{\theta}$) is a computable function (like a neural network), that depends on a set of parameters $\theta$. Converging to the optimal policy is not guaranteed, however, usually, policies converge to close to optimal solutions.

%\snote{todo: remove: There is a concept in trajectory in RL, however, it is defined by sequentaly adding new states, actions and reward, e.g.: Agent execution can be represented with a trajectory e.g.$(s_1,a_1,r_1, \dots, s_n,a_n,r_n)$. Or as this: "In reinforcement learning terminology, a trajectory is the path of the agent through the state space up until the horizon H. The goal of an on-policy algorithm is to maximize the expected reward of the agent over trajectories". It is also refered as HISTORY. Anyways, our trajectory is created independently from it, and norms are checked over our trajectory, so no need for introduction, and I will not mention it in following chapter}

\subsection{From Norms to Rewards}
\label{sub.normsToRewards}

\begin{figure}[h!]
	\includegraphics[width=0.8\columnwidth]{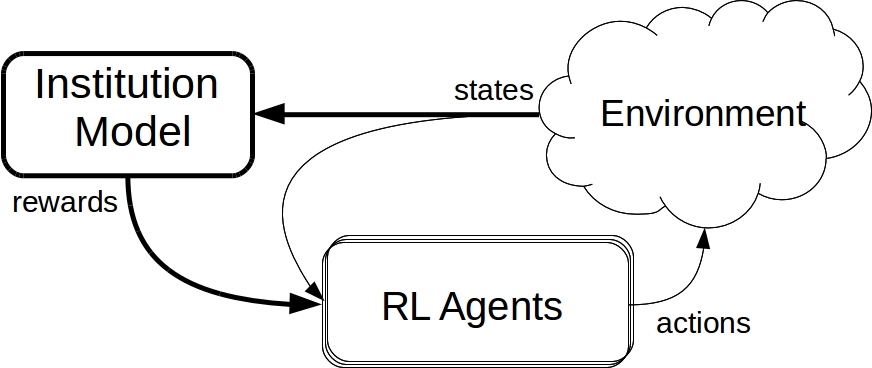}
	\centering
	\caption{RL agents receive feedback signals from the institution. The signals steer agents learning towards normative behavior.}
	%\vspace{-5mm}
	\label{fig:instRewards}
	% \vspace{-5mm}
\end{figure}

The normative semantics in our model is defined using the notion of (see Definition~\ref{def:traje}). A trajectory is constructed from the values in the state space variables which are changing as the agents interact with the environment. This allows us to use norm semantics to evaluate the states of the norms by calculating if they are in fulfilled, neutral, or violated state. This, in turn, can be used to feedback signals regarding the agents' adherence to norms. The schematic view of this approach is shown in Figure~\ref{fig:instRewards}. The `institutional model' box takes only those states of the environment which are used in the norm semantics and evaluates norm states. Naturally, to calculate norm states which have a temporal dimension, a history of states has to be stored. However, state variables are usually shared between various norm semantics, thus the increase of the number of norms usually doesn't mean a proportional increase in the number of stored states. Additionally, as done in our previous research~\cite{tomic2018norms}, instead of storing values of state variables in each time-step, the contiguous interval for which state variables have constant values can be stored, which can significantly speed up evaluation of norm states. This is a common approach used in temporal reasoning, planning and scheduling (e.g.,~\citep{DechterMP91},~\citep{frank2003constraint},~\citep{cesta2008unifying}). Note that the `institutional model' is separated from the concrete RL algorithm and is used only to provide a feedback signal to RL agents. The state space in the RL agents complies with the Markov property (agent action selection depends only on the current state and not on the history of states).

After the states of the norms are evaluated to either fulfilled, violated or neutral, the norm reward function takes the states of norms as an input and returns a feedback signal:
\begin{align*}
\mathcal{F}_{norm} : (NS \times NS) \rightarrow \mathbb{R}
\end{align*}
 %(or $\mathcal{F}_{norm}(ns,ns') = r$, where $ns, ns' \in NS$ and $r \in \mathbb{R}$).
There are $9$ possible transitions of norm states which can be used to signal positive or negative feedback, namely: $(f,f)$, $(f,n)$, $(f,v)$, $(n,f)$, $(n,n)$, $(n,v)$, $(v,f)$, $(v,n)$, $(v,v)$. The $\mathcal{F}_{norm}$ function can be realized in different ways and while its optimal implementation may depend on the underlying RL algorithm, we explore the following more general ideas:

\paragraph{Feedback from Full Adherence} The adherent reward is the main reward or the {\em final goal} for learning normative behavior. Agents receive this reward only when their trajectory is adherent to the institution, i.e., to all institutional norms (see Definition~\ref{def:adheres}). This happens when all transitions are updated to a fulfilled state ($ns'=f$). Algorithm~\ref{alg:adherenceReward} realizes $\mathcal{F}_{norm}$ where the procedure \name{F\_norms} is called at each training step or predefined intervals. The algorithm retrieves, for each norm, its latest state with method \name{GetNS}. If all norms are in the fulfilled state, a reward with the value $1$ is returned. However, this means that an agent has to fulfill all the norms before it receives 'full adherence' reward and does not provide any feedback to norms states separately. Thus this reward is sparse and does not provide helpful intermediate feedback for the credit assignment problem.

\begin{algorithm}[h]
	\small
	\SetEndCharOfAlgoLine{}
	\SetKwFunction{true}{True}
	\SetKwFunction{false}{False}
	\SetKwFunction{adhere}{F\_norms}
	\SetKwFunction{getStateFunct}{GetNS}
	\SetKwFunction{ret}{return}
	\SetKwFunction{isFullfiled}{isF}
	\SetKwInOut{Input}{Input}
	\SetKwInOut{Output}{Output}
	
	\Input
	{
		The current trajectory ([$0,step_t$],$\tau)$); institution $ \inst $; grounding $ \grnd $; norm semantics $ \llbracket \cdot \rrbracket $ %$([0..step_t],\tau), \inst, \grnd, \llbracket \cdot \rrbracket $
	}

	\Output
	{
		1,0 %iff trajectory $([0..step_t],\tau)$ is adherent to $\inst$, 0 otherwise
		%$([0..step_t],\tau) \in \llbracket norm \rrbracket), \forall norm \in Norms$ }
	}

	\SetKwProg{proc}{Procedure}{}{}
	\proc{\adhere{$([0,step_t],\tau), Norms, \grnd, \llbracket \cdot \rrbracket_F $}}
	{
		\ForEach{$norm \in \inst.Norms$}{
			
			$(ns,ns') \leftarrow$ \getStateFunct([$0,step_t$],$\tau)$, $ \grnd $, $ \llbracket norm \rrbracket $) \;
			
%			$\isFullfiled \leftarrow$ \getStateFunct([$0,step_t$],$\tau)$, $ \grnd $, $ \llbracket norm \rrbracket_F $) \;
			\If { $(ns' \neq f)$ } {
				\ret $0$;
			}
		}
		
		\ret $1$ \;
	}
	\caption{Implementation of $\mathcal{F}_{norm}$ for 'Full Adherence Feedback Reward' approach. }
	\label{alg:adherenceReward}
\end{algorithm}

\paragraph{Feedback from Norms States} This approach provides feedback to the agents as they change the states of norms and this is used as a mechanism for reward shaping which indicates {\em partial achievement} of the final goal. The norm reward function assigns fixed numerical values to norm state transitions. For example, transitions from not-fulfilled to a fulfilled norm state $(n,f)$ may be rewarded while transitions to violated states (e.g., $(n,v)$) may be penalized. This is useful for temporal norms since violations of temporal requirements can provide an immediate negative signal. However, this may not be the best approach, since it can lead to local optima, e.g., if among other norms, two temporal norms are present, an agent may break one temporal norm but get the reward from the second one and all other subsequent norms. This is not particularly helpful for the temporal assignment problem and agents can get stuck in local optima. 

\paragraph{Dynamic Feedback} Similarly, as in the previous case, feedback is returned from norm state transitions. However, instead of constant values of norm transitions, they are dependent on each other or controlled by an additional (customizable) function, e.g., ``Stop all future rewards (assign a value of 0) if a violation has happened''. Here, one doesn't want to continue to reward agents trajectories after a violation. This enables agents to learn only from temporally consistent (ordered) normative behaviors. 
%As we will see, this is the most successful approach.

While feedback from norms can be used for reward shaping, and consequently, speed-up learning, we still cannot apply the same abstract norms to novel domains with different agents, behaviors, and objects. 
For example, let's say that $n_1$ is an obligation norm $mustUse(Buyer,Get,Goods)$ and grounding: $\grnd_a = \{Buyer, robot_1\}$, $\grnd_b= \{Get, pick\}$ and $\grnd_o = \{Goods, battery\}$. The policy will learn normative behavior, declared in the semantics of that norm, i.e., that agent \name{robot_1} should execute the particular behavior \name{pick}, and take the particular object \name{battery}. The policy learns on the exact elements represented in the domain state-space. If we change the grounding (using the same norm), e.g., $\grnd_o = \{Goods, box\}$, the agent will not know that it has to \name{pick} a \name{box} now. This means that applying the same norm with different grounding will require retraining. This limitation is addressed below.

\subsection{Abstraction Through Grounding}
\label{sub:abs}

An interesting distinction that is often made in Multi-Agent Systems is whether or not agents know about the organization in which they take part~\citep{boissier2007organization}. The approach described so far is an example of agents not knowing about institutional structure, thus agents are not able to consider normative changes when generating trajectories. Agents can be made aware of the abstract institutional context by explicitly representing norms and grounding as an additional input vector to learning. In such a scenario, agents would have enough information to associate various grounded elements to their normative requirements. The main disadvantage is that this would additionally increase the size of the agents' inputs, hence making learning more difficult. Another, more practical approach, would be to abstract domain elements to the institutional level.

\begin{figure}[t]
	\centering
	\includegraphics[width=1\columnwidth]{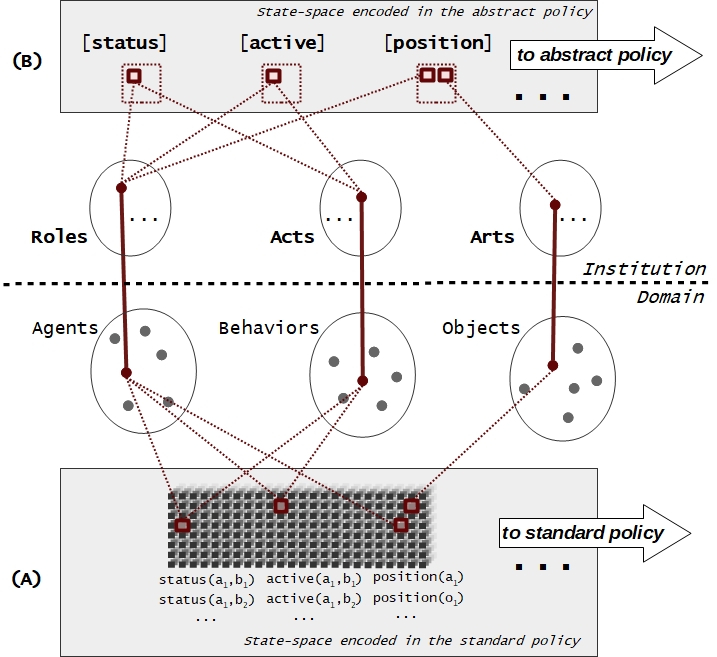}
	\caption{State-space selection for state variables describing institution categories. (A) is the full state space which is usually encoded as an input in the standard policy. (B) represent state-space of abstract institution categories (Roles, Acts, Arts). Grounding (full red line) generates boundary conditions selecting only state-variables describing the state of grounded domain elements.}
	\label{fig:fullBoundary}
\end{figure}

An abstraction of a domain element implies that its meaning is defined in higher, more general categories. Roles, institution actions and artifacts represent such categories, meaningful in social interaction.  
Our institutional model provides a method where an agent's observations and action space can be both abstracted. The equivalence classes such as $[active],[position],[color],[detected],\textit{etc}.$, each represent a type of state-variables characterized by sharing the same attribute or {\em feature}:  `activation', `position', `color', `detection', etc. This is important since it allows us to create an abstracted,  institutional-level state-space, based on such equivalence classes. Changing the grounding of roles, actions or artifacts, may change a particular set of agents, behaviors, and object, but equivalence classes will stay the same and will now describe the state of the new grounded domain elements. Therefore, we can create a stable feature vector based on equivalence classes. Figure~\ref{fig:fullBoundary} illustrates this idea. (A) is full state-space containing all state-variables describing all agents, behaviors, and objects. (B) is institution-level state-space, where states are partitioned by the equivalence classes characterized by the same state-space features. Grounding generates boundary conditions by selecting only state-variables describing the state of grounded domain elements. Thus, the state-space (B) represents states of institution categories (Roles, Acts, Arts). Similarly, outputs of an abstract policy represent institutional actions and are related to concrete behaviors via grounding.

\begin{figure}[t]
	\centering
	\begin{subfigure}[]{0.49\columnwidth}
		\centering
		\includegraphics[width=0.9\columnwidth]{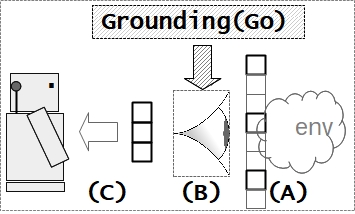}
		\caption{}
		\label{fig:instFocus.input}
	\end{subfigure}
	\begin{subfigure}[]{0.49\columnwidth}
		\centering
		\includegraphics[width=0.9\columnwidth]{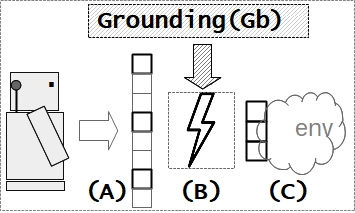}
		\caption{}
		\label{fig:instFocus.output}
	\end{subfigure}
	\caption{(a) Observations: The full state space (A) is filtered through (B) to sub-set of states (C); (b) Actions: The full set of behaviors (A), are reduced (B) to grounded actions (C)}
	\label{fig:instAbs}
\end{figure}

A policy learned in this way would, therefore, be at a higher level of abstraction. We henceforth call such policies ``{\em abstract policies}''. 
Procedural knowledge (knowing how), stored in abstract policies, is learned based on declarative norms semantics defined as relations between (institution-level) categories (roles, actions, artifacts). Grounding selects domain elements that fit into these categories, thus the pattern of activation generated by the abstract policy stays the same but is produced with different (domain-level) agents, behaviors, and objects. This means that in principle, an abstract policy may be applied in every domain where the same (or similar) categories can be identified since they will share the same or similar features over which original policy was learned. Thus, the advantage of abstraction is twofold. First, it can automatically and significantly reduce the size of the observed state-space, addressing the curse of dimensionality problem. Second, an abstract policy can be applied (via grounding) to different agents, behaviors, and objects, and more broadly to different domains, thus achieving a transfer of learning. Figure~\ref{fig:instAbs} shows a schematic view of an abstract policy's inputs and outputs. More about the nature of abstract representation of procedural knowledge and its possible implication to Cognitive Science is discussed in Section~\ref{sec.sub.cogsci}.

Among other things, the ability to re-ground norms to other domains provides a workflow where one can train abstract institutional policies in simulation (adapted to robotics) and then {\em re-ground} them to real robotic systems. The trained policy from simulation then can be updated by learning subtle details with already trained (hence, `safe' from the normative point of view) policy.
 
%======================================================================
% EOF
%======================================================================

\section{Experiments}
\label{sec:experiment}

%\snote{The goal of the experiments}
The goal of this section is to test and evaluate methods described so far. In order to demonstrate the benefits of using normative (social) knowledge and institution abstraction, we choose a simulated environment where it is not so trivial to learn normative behaviors. Thus, we train agents in model-free settings, with relatively long temporal horizons and a large number of states.

The goal of experiment 1 is to empirically prove that agents can indeed learn to behave in accordance with institutional norms. Experiment 2 compares standard policy learning with learning abstracted policies. Here we hypothesize that learning will be significantly faster if the state space is reduced via abstraction. In the subsequent experiment 3, we ground the abstract policy learned in experiment 2 on the new 'factory' domain, demonstrating the transfer of learning. Experiment 4 focuses on reward shaping and we demonstrate its applications on both standard full state-space learning and abstracted learning. In the last experiment 5, we use shaping and abstraction and an additional agent to demonstrate applicability of the method to multi-agent systems. We train two policies in parallel, each controlling the behavior of one agent. 

Videos accompanying this paper can be found at: \url{https://www.youtube.com/watch?v=fzfZD4FyMv4}.

\subsection{Scenario}

The scenario follows the normative behavior specified by the 'Store' institution. Robby is a robot, whose goal is to obtain a pack of batteries. Instead of taking the batteries and immediately exiting the store, Robby is aware of the human institution providing social knowledge of how to behave in buying/selling scenarios. Namely, Robby is aware of the norms: An agent that wants to buy something should pick the desired object and not other objects in the store, and it should pay for the object before leaving the store with it.

The institution is specified as follows (we increase its complexity as we progress through the experiments):

\[
 \textit{Store} = 
 \langle \textit{Arts}_S, \textit{Roles}_S, \textit{Acts}_S, \textit{Norms}_S \rangle
\]
where
{\small
\begin{align*}
  \textit{Roles}_S = \{& \namem{Buyer}\} \\ 
  \textit{Acts}_S = \{& \namem{Pay}, \namem{GetGoods} \} \\
  \textit{Arts}_S = \{& \namem{Goods}, \namem{PayPlace} \} \\
  \textit{Norms}_S = \{& \norm{MustUse}{\stm{Buyer}{GetGoods}{Goods}}\\
                  & \norm{MustAt}{\stm{Buyer}{Pay}{Payplace}} \\
                  & \norm{Before}{\stm{Buyer}{GetGoods}{Goods},\right. \\
                  & \;\;\;\;\;\;\;\;\;\;\;\,\left.\stm{Buyer}{Pay}{PayPlace}} \}
\end{align*}
}

The examples domain include Robby and the set of its behaviors. Robby knows how to take items, pay via wireless money transfer, and open doors. Robby is also capable of moving one step forward and backward, and rotate one step left or right. The domain includes 13 items on sale in the store: battery, drill, ax, screwdrivers etc., and a cash register. The experiments focus only on learning normative policies, and not on particular behaviors (such as grasping an item) thus all robot behaviors are atomic, that is, they are done in one step. In the real case, these could be either other (sub) policies or implemented behaviors. All norm semantics are given previously (see Section~\ref{sub.semantics}).

\subsection{Experiments Setup}

\begin{figure}
	\centering
    \includegraphics[width=1\columnwidth]{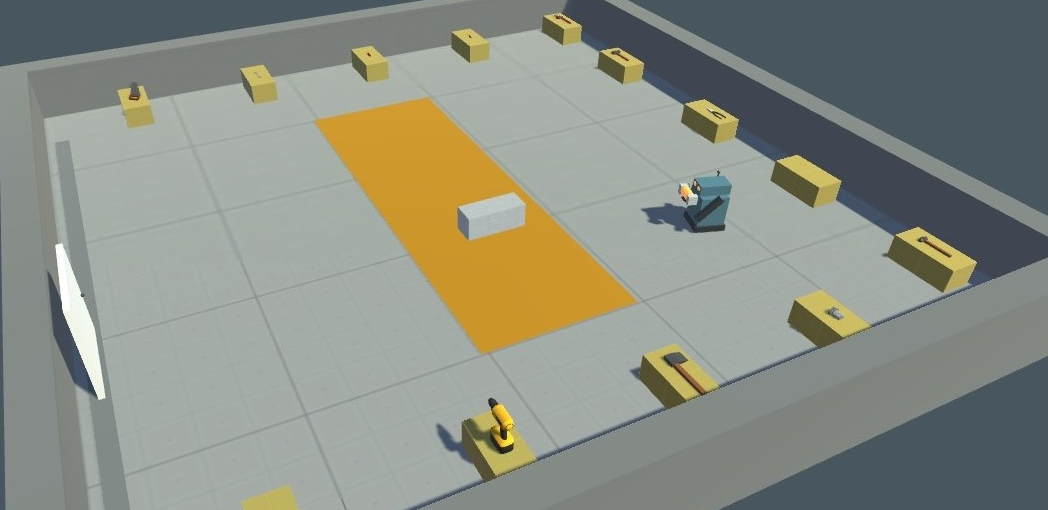}
    \caption{Simulated Store Environment}
    	%, containing the agent Robby, shelves with various items, payment-register (white box in the middle) and the door}
    %\vspace{-5mm}
    \label{fig:simEnv}
    % \vspace{-5mm}
\end{figure}

\paragraph{Software}
Model-free RL requires a substantial amount of training data, which we generate via simulation. In this paper we used a game engine~\citep{misc:unity} to simulate a simplified store domain (see Figure~\ref{fig:simEnv}). The state of the art algorithm Policy Proximal Optimization (PPO)~\citep{schulman2017proximal} is used to train simulated agents. This is done with the help of the Machine Learning toolkit~\cite{juliani2018unity} (v0.6), which includes a TensorFlow implementation of PPO and a bridge between simulation data and the RL algorithm. For all reported experiments we used maximum 2000 steps per episode. Environment and agents restart if full adherence is achieved or if the simulation step reaches the maximum count. In addition to the rewards discussed earlier, the agent receives small negative feedback ($1.0e-4$) for each simulation step so as to make agents finish the episode as soon as possible. We used the same hyper-parameters in all experiments and some of them are shown in Table~\ref{tab:my-table}, where newly introduced $beta$ is the entropy regularization parameter (increasing it will influence policy to take more random actions). Time\_horizon is the length of generated trajectory used for gradient update and there are 3 layers of 256 training units. Parameters were chosen as a result of pilot tests (not reported here).

\begin{table}[h]
	\centering
	\resizebox{\columnwidth}{!}{%
		\begin{tabular}{llllllll}
			\hline
			beta                     & gamma                    & lambd                    & learning\_rate           & num\_layers           & hidden\_units           & time\_horizon            & batch\_size \\ \hline
			\multicolumn{1}{c}{6e-3} & \multicolumn{1}{c}{0.99} & \multicolumn{1}{c}{0.95} & \multicolumn{1}{c}{3e-4} & \multicolumn{1}{c}{3} & \multicolumn{1}{c}{256} & \multicolumn{1}{c}{1024} & 1024        \\ \hline
		\end{tabular}%
	}
	\caption{Hyper-parameters used for training in the experiments}
	\label{tab:my-table}
\end{table}

\begin{figure}
	\centering
	\begin{subfigure}[]{0.9\columnwidth}
		\includegraphics[width=1\columnwidth]{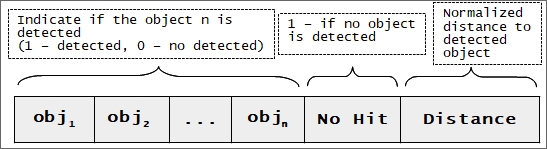}
		\caption{}
		\label{fig:input.standard}
	\end{subfigure}
	\begin{subfigure}[]{0.9\columnwidth}
		\includegraphics[width=1\columnwidth]{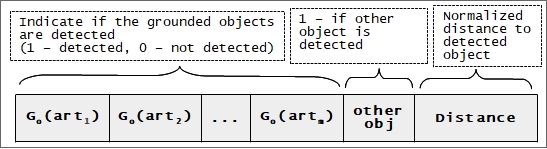}
		\caption{}
		\label{fig:input.abstract}
	\end{subfigure}
	\caption{Vector input from one ray in the robot sensor in institution focus approach}
	\label{fig:instFocus}
\end{figure}

Robots observe the state of the environment through a sensor consisting of 7 rays positioned at the center of the agent, pointing in different directions. Each ray returns a vector which encodes information about detected objects and its normalized distance (see Figure~\ref{fig:input.standard}). Additional inputs are: velocity vector of the agent, information about whether behaviors are already executed, and state variables `near', indicating whether an object is near the agent and `has', indicating whether the agent holds an object or not. The input vector size is 162 elements.
However, the abstract policy approach encodes only states regarding grounded objects, while states regarding other domain elements are encoded in only one vector element merely as an indication of their existence reducing the input vector to 39 elements (see Figure~\ref{fig:input.abstract}).

\paragraph{Hardware} 
The simulations are executed on a desktop computer with the following configuration: Intel Core i7 4790K CPU @ 4GHz (x64), 16GB DDR3 RAM, Nvidia GTX970 (4GB GDDR5). 

\subsection{Experiment 1: Proof of Concept}
\label{sub.exp1}

The goal of this experiment is to learn the normative behavior specified by the Store institution. The experiment is executed in a simplified Store environment with 3 items, using only adherent reward based on
algorithm~\ref{alg:adherenceReward}. Groundings are given as follows: $\grnd_a = \{Buyer, Robby\}$, $\grnd_b = \{GetGoods, pick\}$, $\grnd_b = \{Pay, transfer\}$ and $\grnd_o = \{Goods, battery\}$.

\begin{figure}[h]
	\centering
	\includegraphics[width=0.8\columnwidth]{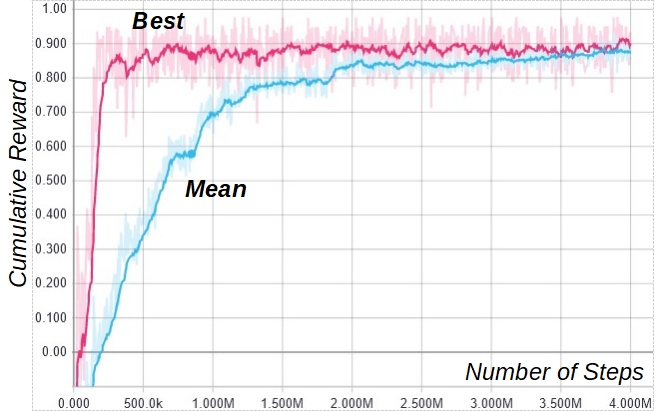}
	\caption{Experiment 1: Robby learning using only adherent reward in simplified environment}
	%\vspace{-5mm}
	\label{fig:exp1}
	% \vspace{-5mm}
\end{figure}

Results are shown on Figure~\ref{fig:exp1}. The $x$ axis shows the number of steps, while the $y$ axis shows the amount of accumulated reward. `Mean' line represents the learning curve averaged over 10 independent training trials, where for each training trial data is collected over 16 parallel simulations of the environment. Figure~\ref{fig:exp1} also shows the best learning curve (out of 10 trials), which reached near optimal policy in the minimum number of steps. 

\subsection{Experiment 2: Learning Abstract Policies}
\label{sub.exp2}

\begin{figure}[h]
	\centering
	\begin{subfigure}[]{0.8\columnwidth}
		\includegraphics[width=1\columnwidth]{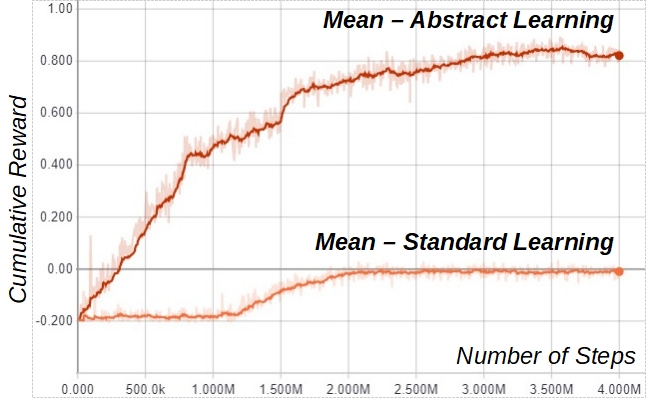}
		\caption{}
		\label{fig:exp2_means}
	\end{subfigure}
	\begin{subfigure}[]{0.8\columnwidth}
		\includegraphics[width=1\columnwidth]{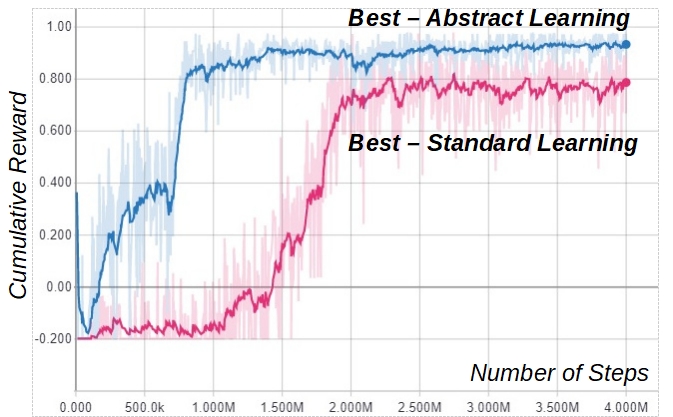}
		\caption{}
		\label{fig:exp2_bests}
	\end{subfigure}
	\caption{Experiment 2: Comparison between standard learning versus abstract learning: (a) means over 10 trials (b) best results}
		%, both using only adherent reward, where 4 trials are executed for 4M steps. In (a) Robby didn't learn normative policy, while in (b) it has learned near-optimal policy}
	\label{fig:exp2}
	%\vspace{-5mm}
\end{figure}

In this experiment, we turn our attention to abstracted policy learning. We increase the complexity of the Store domain, by adding $13$ different store items. For training, we use only adherent reward, as described in section~\ref{sub.normsToRewards}, in two different settings: (A) Standard learning (B) Abstract Learning. The hypothesis is that the learning will be significantly faster in setting (B) since the agent state-space is significantly reduced. Results confirm our hypothesis. Figure~\ref{fig:exp2_means} shows two learning curves for settings (A) and (B), each of them representing a mean of 10 different execution trials. In setting (A) the RL agent didn't manage to learn a normative policy in almost all the trials, wherein in settings (B) almost all training trials managed to converge to near-optimum behavior. The best and only successful trial in standard learning (A) is compared to the best trial in abstract learning (B) (see Figure~\ref{fig:exp2_bests}). Note that the abstract policy is applicable to a whole category of inputs -- a subset of equivalence classes of state-variables -- while the standard policy is applicable only to one particular (fixed) grounding. In that sense, these policies are not the same, and the only reason we can compare them is that we have used the same grounding for both of them. 

The following experiment is concerned with demonstrating and understanding the quality of an abstract policy when used with a grounding other than that used for learning.

\subsection{Experiment 3: Transfer of Learning}
\label{sub.exp3}

\begin{figure}[h]
	\centering
	\includegraphics[width=0.9\columnwidth]{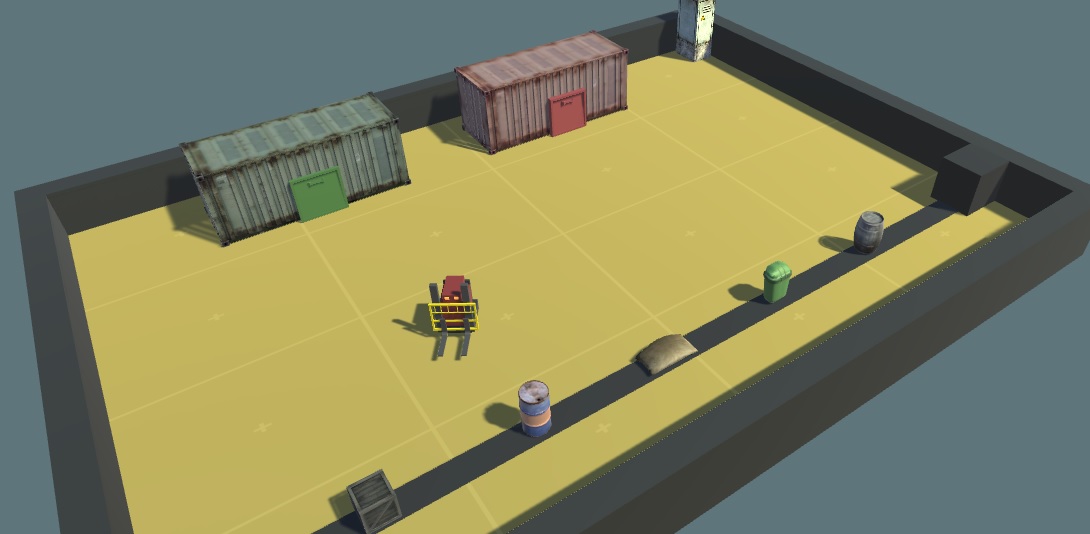}
	\caption{Factory-yard environment}
	%\vspace{-5mm}
	\label{fig:factory}
	% \vspace{-5mm}
\end{figure}

In section~\ref{sub:abs}, we hypothesized that a policy learned at the level of abstraction of institution can be used across boundary conditions and domains. Thus, the goal of this experiment is to apply the same set of norms in a novel domain. In scenario (A) we test a new grounding in the same domain: $\grnd_o = \{Goods, drill\}$. We expect that the agent will know to buy the drill instead of the battery since now the drill is abstracted to 'Goods'. It is important to stress that while humans may assign different names to roles, actions or artifacts over different domains, the pattern of agents behaviors may still have the same semantics over such domains. Scenario (B) demonstrates this concept, where we use a factory-yard domain (see Figure~\ref{fig:factory}). 
A robot named Forky is required to sort out different items by locating them on a conveyor belt, lifting them and bringing them to a container's hatch for disposal. We ground the Store institution to the new domain as follows: $\grnd_a = \{Buyer, Forky\}$, $\grnd_b = \{GetGoods, lift\}$, $\grnd_b = \{Pay, leave\}$ and $\grnd_o = \{Goods, box1\}$. Forky is twice as slow than Robby, the size of the environment is increased, and the spatial layout is somewhat different. Additionally, items that Forky is required to lift are moving, and sometimes they are not in the environment at all (they appear on the conveyor belt), which can additionally confuse the agent.
% since it introduce the need of 'waiting' and 'time' which are not modeled in the agents state-space. 
In the trials, we distinguish between: (B1) - directly applying the abstract policy learned in the Store scenario; (B2) where we continue training of the transferred policy in the new domain for an additional 200k steps; and (B3) where we train the agent from scratch.

\begin{figure}[h]
	\centering
	\includegraphics[width=0.9\columnwidth]{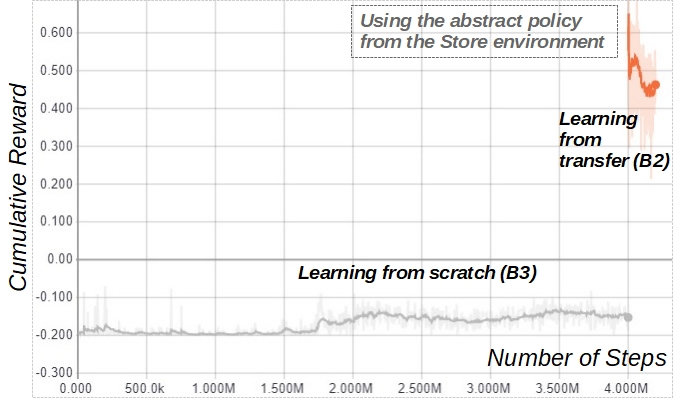}
	\caption{Experiment 3: Results averaged over 10 different trials of learning from abstract Store policy (B2) compared to learning from scratch(B3). }
	%\vspace{-5mm}
	\label{fig:exp3}
	% \vspace{-5mm}
\end{figure}

Results show that in the trail (A), Robby successfully locates the drill (instead of the battery), picks it up and goes to the cash register to pay. In scenario (B), with the original store policy (case B1), Forky manages to search for and navigate to the grounded factory item (box), lift it, and in most of the episodes manages to reach the hatch to dispose of the item. Figure~\ref{fig:exp3} shows continuing training the policy (B2) which did not improve agent behavior, while learning from scratch (B3) did not manage to achieve any significant results within 4M steps. Results are averaged over 10 different trials. 
Interestingly, the cumulative reward in B2 started to decline which shows that learning in the factory-yard is very difficult (given its dynamics), whereas desired behavior is easily achieved via a transfer of policies learned in a simpler environment. Simple environments allow agents to base their learning on exploiting salient features in the environment. In the factory-yard domain, there are disturbing, dynamic elements that have nothing to do with norms and it is much harder to capture salient features leading to norm adherence. Hence, the experiment demonstrates that it is much easier to capture such features in an abstract policy learned in a simpler environment and then applying it in ``normatively-equivalent'' dynamic environment. 
The ability to transfer knowledge is also important since it provides a way of learning abstract policies that can capture salient features responsible for norm adherence in simulations and then ground them to real-world domains with complex dynamics. Real robotic agents may continue to learn (subtle) differences between simulation and the real-world in a safer and faster way than starting from scratch.

\subsection{Experiment 4: Normative Feedback}
\label{sub.exp4}
 
Figure~\ref{fig:exp2} shows that the policy was not able to converge (in almost all of the trials) in a complex state-space without intermediate feedback. However, the information about norm states can help the learning agent. The goal of this experiment is to test reward based on feedback from individual norm states and 
to assess whether it can more effectively/quickly guide an agent towards a normative policy. The temporal norms are hardest to learn since they directly introduce the (temporal) credit assignment problem. We make the problem even more challenging (compared to experiment 2) by adding an additional temporal norm where now the buyer, after paying, is expected to exit the store through the door. The institution is extended with another action $\namem{Exit}$ and artifact $\namem{ExitPlace}$ and the following norms:

{\small
\begin{align*}
    & \norm{MustAt}{\stm{Buyer}{Exit}{ExitPlace}} \\
	& \norm{Before}{\stm{Buyer}{Pay}{PayPlace},\right. \\
    & \;\;\;\;\;\;\;\;\;\;\;\,\left.\stm{Buyer}{Exit}{ExitPlace}}
\end{align*}
}

The hypothesis here is that Robby should learn the policy in a more complex environment even without institution abstraction, however, the abstraction will lead to even faster learning. We test two similar approaches designed to mitigate the temporal assignment problem:

\textit{(A.) Stop rewards after violation.} Assign the same amount of reward to each norm: $1.0$ / \name{(number \; of \; norms)}, for transition $(n,f)$. However, stop rewards if any of the norms are violated. This avoids local optima. The idea here is to steer the agents towards temporally-consistent behaviors. Note that adherent reward of value $1.0$ is still given if all norms are satisfied, making the maximum possible reward a little less than $2.0$ (taking into account small negative feedback from the environment at every step).

\textit{(B.) Restart after the violation. } The same shaping function as (A), with the difference that an episode is finished (restarted) immediately after the violation. Agents do receive the negative feedback equal to the sum of environmental negative feedback from each step, which an agent would get anyways before the end of an episode. The idea is to avoid any future (ill-informative) updates of the policy since the norm reward function will not give any feedback even when other norms are achieved separately (as in case A). In addition, this may shorten overall training time, since when a norm is violated the agent doesn't have to wait until the end of the episode.

\begin{figure}
	\centering
	\begin{subfigure}[]{0.8\columnwidth}
		\includegraphics[width=1\columnwidth]{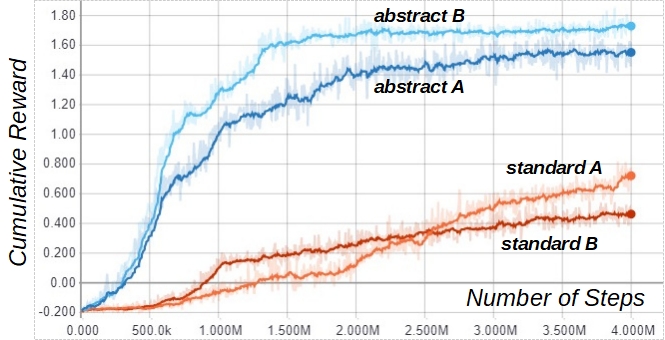}
		\caption{}
		\label{fig:exp4_means}
	\end{subfigure}
	\begin{subfigure}[]{0.8\columnwidth}
		\includegraphics[width=1\columnwidth]{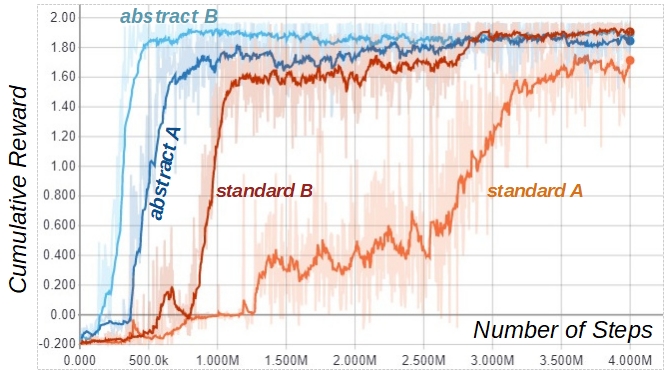}
		\caption{}
		\label{fig:exp4_bests}
	\end{subfigure}
	\caption{Experiment 4: Rewards shaping. Two implementations (A) and (B) of a reward function applied for (a) standard and (b) abstract learning}
	\label{fig:exp4}
	%\vspace{-5mm}
\end{figure}

Figure~\ref{fig:exp4_means} shows means over 10 training trials for both shaping approaches (A) and (B) applied to standard (full-state space) learning and abstracted learning. Both approaches are similar, where slightly better results are achieved with the approach (B). While, experiment 2 (see Figure~\ref{fig:exp2}) shows that standard learning without shaping is unsuccessful in 4M episodes, in this experiment with additional temporal requirements, results are significantly improved and the learning curve is constantly increasing. Still, 4M steps are not sufficient for all trials to reach their near-optimal policies. Figure~\ref{fig:exp4_bests} shows the best results in all tested approaches.

\subsection{Experiment 5: Multiple Agents}

No specific changes have to be introduced in order to train multiple agents. While it is possible to learn one abstract policy to guide all the agents in the institution, in this experiment each agent learns its own abstract policy. The rewards are distributed in the following way. The acting of all agents together creates a single domain trajectory which can be either adherent to an institution or not, thus the final reward from the full adherence is given to each agent that is grounded by the institutional roles. Norms of particular agents depend on their role in the grounding, thus each agent gets a feedback from the norm shaping function for fulfilling the norms that are relevant to it. Some norms are defined over more than one role: in such cases, agents have to cooperate to receive a feedback reward.

In this experiment we use an additional robotic agent named 'Kobby' who is capable only of navigating, receiving and accepting/declining payments from the nearby agents. The institution now includes another role \name{Seller}, and action $\namem{ReceivePayment}$, with additional norms indicating that the seller has to receive payment at the place of payment and a temporal norm 'equals' making sure that agents will synchronize their behaviors, since its semantics specifies that paying and receiving payment has to happen at the same time: 

{\small
\begin{align*}
  & \norm{MustAt}{\stm{Seller}{ReceivePayment}{PayPlace}} \\
   & \norm{Equals}{\stm{Buyer}{Pay}{PayPlace},\right. \\
                  & \;\;\;\;\;\;\;\;\;\;\;\,\left.\stm{Seller}{ReceivePayment}{PayPlace}}
\end{align*}
}

\begin{figure}[h]
	\centering
	\begin{subfigure}[]{0.8\columnwidth}
		\includegraphics[width=1\columnwidth]{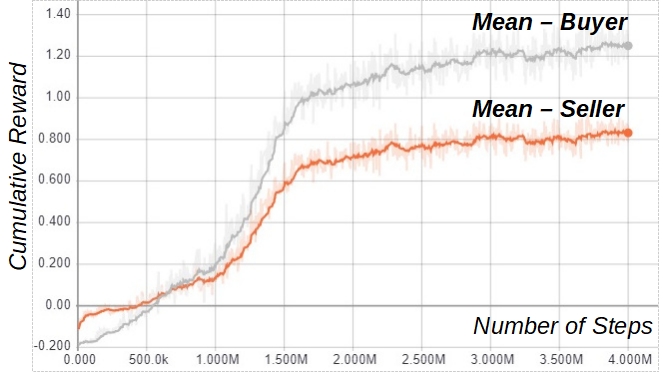}
		\caption{}
		\label{fig:exp5_means}
	\end{subfigure}
	\begin{subfigure}[]{0.8\columnwidth}
		\includegraphics[width=1\columnwidth]{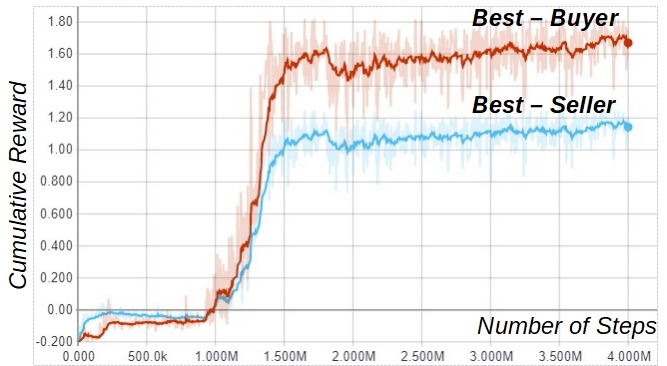}
		\caption{}
		\label{fig:exp5_bests}
	\end{subfigure}
	\caption{Experiment 5: Two policies (buyer and seller) are trained in parallel: (a) Mean over 10 trials (b) Best out of 10 trials}
	\label{fig:exp5}
	%\vspace{-5mm}
\end{figure}

Similarly as the temporal semantics of `\name{before}', the fulfillment semantics of `\name{equals}' can be defined by changing relation between $t_1$ and $t_2$ to $t_1 = t_2$. Note that the norm is defined between triples involving different roles. This means that agents grounded to the role of 'Seller' and 'Buyer' have to cooperate to achieve adherence, i.e., `Paying' of one agent should be at the same time as `ReceivingPayment' of another agent.
Figure~\ref{fig:exp5} shows that both agents have learned their normative policies. The 'Buyer' has to fulfill more norms that the 'Seller', which results in the difference in cumulative reward. Both learning trials converge to near optimal policies. 

\subsection{Summary of Results}

The overview of the experiments and the main results are shown in Table~\ref{tab:overview}. Results reveal that learning normative behavior using only adherent feedback at the end of the episode, is possible (Experiment 1), but it is not convenient in more complex state-spaces (Experiment 2). Much better results are achieved when the state-space is simplified with respect to grounding when learning abstract policies (Experiment 2). Abstraction also enables applying such policies to novel domains, even when the policy in a novel domain is very hard to, or cannot be learned from scratch (Experiment 3). Furthermore, results show that normative reward feedback is highly useful to steer the learning algorithm towards successful policies, even in complex environments (Experiment 4). However, the best results are achieved when combining both approaches together: intermediate normative feedback and (institution) abstraction (Experiment 4). Finally, it is shown that, in the same manner, it is possible to learn policies for coordinating multiple agents (Experiment 5).

\begin{table}[]
	\centering
	\resizebox{\columnwidth}{!}{%
		\begin{tabular}{@{}cccccc@{}}
			\toprule
			& Exp1 & Exp2 & Exp3 & Exp4 & Exp5 \\ \midrule
			\begin{tabular}[c]{@{}c@{}}State-Space\\ Complexity\end{tabular} & Low & High & High & High & High \\ \midrule
			\begin{tabular}[c]{@{}c@{}}Temporal\\ Complexity\end{tabular} & Low & Low & Low & High & High \\ \midrule
			\begin{tabular}[c]{@{}c@{}}Standard\\ Learning\end{tabular} & Success & Failed & - & - & - \\ \midrule
			\begin{tabular}[c]{@{}c@{}}Standard\\ Learning\\ (Shaping)\end{tabular} & - & - & - & \begin{tabular}[c]{@{}c@{}}Success\\(needs more data)\end{tabular} & - \\ \midrule
			\begin{tabular}[c]{@{}c@{}}Abstract\\ Learning\end{tabular} & - & Success & \begin{tabular}[c]{@{}c@{}}Failed (Learning)\\ Success (Transfer)\end{tabular} & - & - \\ \midrule
			\begin{tabular}[c]{@{}c@{}}Abstract\\ Learning\\ (Shaping)\end{tabular} & - & - & - & \begin{tabular}[c]{@{}c@{}}Success\\ (Best)\end{tabular} & \begin{tabular}[c]{@{}c@{}}Success\\ (Multiple Agents)\end{tabular} \\ \bottomrule
		\end{tabular}%
	}
	\caption{Summary of the Results}
	\label{tab:overview}
\end{table}

%======================================================================
% EOF
%======================================================================

%======================================================================
\section{Discussion and Future Work}
%======================================================================
\label{sec.discussion}

Our current framework provides a way to model a set of norms relevant to a given social context. Agents can learn to behave and cooperate with respect to their assigned role in a way predictable by humans. Still, focusing on only one institution at the time can be limiting. In real-world examples, humans often follow norms of multiple institutions concurrently. For example, normative behavior in a store also includes adhering to other generally-accepted behavioral norms, like not bumping into others, not making excessive noise, etc. 
Also, knowing that an institution, e.g., a gas station, is subordinate of a more general category of paying/trading institution, an agent can assume that it has to pay for the gas, without explicitly specifying it in a gas-station institution. Norms from different institutions on different hierarchical levels should be combined together to produce a full set of desired normative behaviors. The study of how different institutions can be related to each other, and how to exploit these relations in verification, planning, and learning, should be investigated further. Naturally, norms concern robot safety, and while we plan to test transfer of normative (safe) policies to the real world via grounding, safety issues go deeper than that~\cite{amodei2016concrete}. One of the important questions that need answering is, can humans identify the `critical mass' of norms that would ensure expected social behavior in all circumstances?

This paper focuses on demonstrating how our normative framework can be used to create policies that generate normative behaviors. Still, the described method suggest possibilities beyond those we have described. For example, imagine an agent exploring an environment. Such an agent can be rewarded for any trajectory for which an admissible grounding can be found and which is adherent to institutional norms. Since a much larger subset of trajectories can be used for learning, such an algorithm should be more data-efficient, and learn from fewer examples. Another unexplored possibility would be to realize curriculum learning~\cite{bengio2009curriculum} by gradually increasing the difficulty of the problem by changing the grounding. Similarly, instead of randomizing items in our simulated environment for each training episode, we could have randomized grounding. An interesting direction is to investigate the nature of the abstract knowledge representation in the broader context of AI and Cognitive science. Some brief discussions are covered below.

\subsection{Connections to Cognitive Science}
\label{sec.sub.cogsci}

A policy is a sensory-motor (perception-action) map. Standard policies represent all sensory inputs and actions outputs, while abstract policies have abstract, that is, inputs/outputs related to categories. Abstract policies store procedural knowledge for achieving declarative semantics specified on an abstract (categorical) level. In psychology and cognitive science, a pattern of behaviors that organize categories of information and relationships between them is known as `{\em schema}' (e.g.~\citep{dimaggio1997culture}).  
The term `schema' was initially introduced in psychology by \citet{bartlett1932remembering}. The paper proposes that human knowledge is stored in underlying mental structures that represent a generic knowledge about the world. Since then, many other terms are used to describe schema, such as: `frame', or `script'. They are also referred to as a `mental' models or representations, `concepts', etc.

% our work = schemas
As described by ~\citet{hampson1996understanding}, schema (or plural `schemata'), ``store both declarative (`what') and procedural (`how') information, where declarative knowledge is knowing facts, knowing that something is the case, while procedural knowledge is knowing how to do something''. Exactly this is the ontology modeled in our framework (see~\ref{sub.ontology}).
Furthermore, ~\citet{alba1983memory} describe four important processes relevant to schemata: (1) \emph{choosing incoming stimuli}, which can guide attention and focus only on relevant stimuli; (2) \emph{abstraction}, which stores the meaning without the details (its original content); (3) \emph{interpretation} of the new information by association to previously-stored knowledge; and (4) \emph{integration} of those processes into a memory. It can be argued that we have made a step towards achieving these processes in our computational framework, thus creating a schematic knowledge representation. Choosing incoming stimuli is done via grounding. The knowledge stored in abstract policies does not contain information about a particular  domain, rather it stores information regarding relations between (social) categories. As such, it can be used to interpret information about new domain elements depending on previously-stored relations, by fitting them into corresponding (social) categories. By exploiting properties (2) and (3) we were able to achieve the transfer of learning (see Section~\ref{sub.exp3}). 

Early work in computer science, focusing on creating `frames' in machines, was conducted by ~\citet{minsky1975patrickFIX}. He argued that humans are using stored knowledge about the world to accomplish many of the processes that the framework was attempting to emulate. Based on this work, new theories about mental representation emerged in the field of cognitive psychology (e.g.,~\citep{rumelhart19801980}). The importance of schematic knowledge representation is explained in the book by ~\citet{thagard2005mind}, stating that people have ``mental processes that operate by means of mental representations for the implementation of thinking and action''. 
Indeed, schemata are the foundation of numerous theories regarding modeling human cognition, concepts formation, language development and comprehension, culture, representational theories of mind, etc. Being able to realize explicit schematic (conceptual) knowledge through our computational model, may have further influence on the development of such theories.
While this paper is focused on representing social normative knowledge, the knowledge about other properties of the world is represented with other types of schemata, for instance, self-schema, knowledge about oneself based on past and grounded in present experiences; object-schema, knowledge about different categories of objects, their function, and structure, etc. How to design and combine different types of schemata, and use it for robots is an interesting and promising area for future investigation.

The area of 'Grounded Cognition'~\citep{barsalou2008grounded}, states that {``conceptual representation underlying knowledge is grounded in sensory and motor systems''} and perception-action links are used ``as a common base of simple behavior as well as complex cognitive and social skills''. 
One of the main questions to answer in this filed is: how abstract concepts such as `democracy'~\citep{cangelosi2018review} can be formed from the sensory-motor experience? A possible solution to this problem is proposed by~\citet{barsalou1999perceptions}, where such abstract concepts can be created through mental simulations and {\em conceptual combinations}. Similarly,  \citet{svensson2004making} argue based on evidence from a range of disciplines, that higher-level concepts and cognitive processes are based on {\em simulations} of sensory-motor processes. Robots are often used for the demonstration of such theories~\citep{pezzulo2013computational}. A review of computational modeling approaches concerned with learning abstract concepts in embodied agents (robots), is given by ~\citet{cangelosi2018review}. 
%Whether such conceptual combinations and explorations can be related to \textit{`Mentalese'}, a hypothetical language of thought~\citep{fodor1975language}, or not, should be further explored in the multidisciplinary research effort.
%
Our model allows us to represent some of the abstract concepts that can be specified as an institution. For instance, it would be possible to specify an `election' institution, including roles such as leader and voter, and actions that count-as voting --- inserting a paper into a box. However, defining more complex concepts (e.g., `democracy'), would, probably, require relating multiple institutions.

At the current stage of our work, we are not combining abstract knowledge into more complex structures. This seems like an important direction for future study. The question of how conceptual knowledge may be combined via a combination of abstract policies is discussed below (see Section~\ref{sec.sub.futurework}). 

\subsection{Connections to Cognitive Linguistics}

Cognitive and linguistic theories are linked together in the multidisciplinary area of cognitive linguistics. The knowledge representation is often necessary for explaining the acquisition, comprehension, and use of language and its relation to meaning. Jean Piaget's theories~\citep{mcleod2009jean} suggest two processes, central for child development: accommodation and assimilation. Former describes creating new or changing old schemes when a certain situation doesn't make sense, while the latter is understanding the world through already existing schemata. He argues that children first have to develop a mental representation of the world (schemata) and then base their language on such representations.
Our framework is not focused on linguistic representation. Language has a deeper and more complex structure than our current framework can express. Still, some interesting connections exist and deserve a brief discussion. Norms are part of human language, and our work on how to make robots `understand' norms in terms of execution, may prove to be similar to how language is understood in general (or at least some of its parts). The fact that language is a social phenomenon, that its meaning depends on context, and given that both triples, role-action-artifacts, and subject-predicate-object, abstract agents interactions, indicates that a computational model based on related principles of categorical abstraction may be created. 
%
%Searle - language is institution?.
%
For instance, a simple sentence like ``Bring me a glass of water'', can be grounded by the same abstract knowledge representation used in `Store' institution (see Section~\ref{sec:experiment}), where a robot should now go and pick up (GetGoods) the glass of water (Goods) and bring it to a human (Register). Similarly, the same knowledge can be grounded to a sentence with the same meaning in other languages. Furthermore, a simple sentence such as ``John opens the door'' has a certain underlying (abstract) semantics, which can be captured with our model. Humans know how to ground abstract knowledge to language symbols in the syntactically right way. The wrong grounding would produce word order that would lead to ill-formed sentences or categorical errors, which cannot make sense in terms of execution. For example, switching the grounding of the agent with the grounding of the object, in the same sentence, would result in the meaning of the form ``Door opened John''. Moreover, a sentence using words such as `who', `what', etc., indicating the lack of grounding. For instance ``Who opens the door?'', lacks the subject or (agent) grounding. Finally, the grounding depends on object affordances (see~\citep{tomic2018norms}), thus syntactically correct sentence with not-admissible grounding will not have sense, e.g., ``John jumps the door''.  
%(or the famous Chomsky sentence ``Colorless green ideas sleep furiously'').
The same abstract knowledge may be grounding to many different symbols, thus forming many different sentences with the same meaning. 
Naturally, the question arises: is such or similar abstract knowledge representation a step towards realizing the universal property for all languages (language universals)? 

Finally, `metaphors' are often used in linguistic theories: ``The essence of metaphor is understanding one kind of thing in terms of another''\citep{johnson2003metaphors}. According to this definition, metaphors could be explained with `interpretation' property (see Section~\ref{sec.sub.cogsci} or ~\citet{alba1983memory}) of abstract knowledge (schema). One of the key questions is ``whether these metaphors simply reflect linguistic convention or whether they actually represent how people think''~\citep{murphy1997reasons}.

\subsection{Future Work}
\label{sec.sub.futurework}

We prioritize on the investigation of what is possibilities to achieve with abstract policies in computer simulation while applying and testing obtained results on real robots. 

\textit{Exploring Grounding.}
%In Section~\ref{sec.sub.cogsci} we have discussed the idea of {\em conceptual combinations}~\citep{barsalou2008grounded} as a means to create more complex concepts. A comparable idea may be explored in a computer simulation. 
%
%The norms in our framework are defined as qualifiers over triples which, through semantics, characterize different types of relations between categories of roles, actions, and artifacts. As part of an institution, together they specify {\em what} is regarded as normative behavior in a given situation.
%
While in this paper we have demonstrated learning of abstract policies based on all institution norms, it is also possible to learn each norm {\em independently} and {\em separately} and store them in separate `primitive' abstract policies. Such policies could then be {\em combined} to achieve normative behavior as specified by the institution, by finding the grounding that produces an adherent trajectory. Furthermore, primitive abstract policies can be used outside the scope of their normative meaning. Each of them would contain procedural knowledge of how to achieve one particular relation described by the norm qualifier (e.g., prepositions) regarding temporal relations (after, before, during, overlap, etc.), spatial (at, near, on, below, etc.), but also other relation of interest. An agent grounded by a primitive policy would act towards  `achieving' it. For instance, if a policy that captures concept `at', grounds an agent and an object representing a football field, the agent will act towards being {\em at} that football field. Changing the grounded object would make the agent move to another area and activating another primitive policy will make the agent act towards `achieving' the new policy. Thus, it would be possible to explore the world, by {\em exploring grounding} of such policies, which can be further used for learning or planning. This possibility seems interesting to explore and would require learning a set of primitive abstract policies over diverse examples. Then another policy will output a vector encoding possible groundings and possible primitive policies to be activated. This will allow learning groundings of primitive policies in order to generate agents behaviors that maximize certain reward. Also, such a design may make learning on the level of granularity more meaningful from the human perspective.

\textit{Hierarchical/Recursive Grounding.}
The described hypothetical process of learning grounding of `primitive' abstract policies and combining them into more complex representations, would still have a `flat' structure. A complementary idea in this direction of research is to investigate grounding of `primitive' policies on other such policies. In an early investigation, we were able to define norms in terms of other norms, where we have defined the `before' norm over two other `must' norms. The agent has successfully learned to fulfill both `must' norm in correct sequential order. Such hierarchical grounding was implicit, and future research effort should make it explicit. Being able to hierarchically ground abstract policies would mean that simple concepts could be combined in hierarchical structures, which should enable agents to understand more complex forms of norms, and explore the world on different levels of granularity, just by exploring grounding.

\textit{Grounding and Transfer of Knowledge.} 
At the current level of development of our model, whether an abstract policy can be applied in a given domain, or not, can be decided by a human knowledge engineer. However, an interesting question needs answering is: how should a computational system know to automatically ground an abstract policy in novel domains? In  Section~\ref{sub:abs} we explain that the abstract policy can be (in principle) grounded in  domains that shares the same features as the categories over which abstract policy is originally learned. However, in practice, the success of a transfer may depend on other morphological/isomorphic (syntactic/semantics) similarities. Related ideas are suggested in the area of analogy reasoning~\citep{sep-reasoning-analogy}~\citep{gentner1983structure}, and they may serve as a source of inspiration for future research. Researches in this area were able to explain how a mapping of a domain to other domains may result in creativity and some forms of reasoning~\citep{gust2008analogical}.

In this section, we have briefly discussed the nature of abstract knowledge representation in our framework, some ideas for future work and its possible implications to other areas. Interestingly, in this view, all of the discussed areas may have something in common: the grounding of abstract knowledge representation. A multi-disciplinary research effort may be needed for a unifying cognitive model where the grounding of abstract knowledge plays a central role.

%======================================================================
% EOF
%======================================================================

%======================================================================
\section{Conclusions}
%======================================================================
\label{sec.discussion}

Future robots will need to follow human social norms in order to be useful and accepted in human society. On the other hand, recent successes in reinforcement learning techniques may further speed up the development of robotics. In this paper, we propose a method to bring social norms to reinforcement learning algorithms together through our institutional framework. We are able to: (1) provide a way to intuitively encode social knowledge (through norms); (2) guide learning towards normative behaviors (through an automatic norm reward system); and (3) achieve a transfer of learning by abstracting policies; Finally, (4) the method is not dependent on a particular RL algorithm (although it was tested with a particular RL algorithm). We show how our approach can be seen as a means to achieve abstract procedural knowledge representation and we discuss its implication to cognitive science. 

%======================================================================
% EOF
%======================================================================

% BibTeX users please use one of
%\bibliographystyle{spbasic}      % basic style, author-year citations
%\bibliographystyle{spmpsci}      % mathematics and physical sciences
%\bibliographystyle{spphys}       % APS-like style for physics
%\bibliography{}   % name your BibTeX data base

%\bibliographystyle{spbasic} % this works with citep and citet from natbib package
%\bibliographystyle{spmpsci}      % basic style, author-year citations

\bibliographystyle{IEEEtranN}
\bibliography{biblioST}   % name your BibTeX data base

% Non-BibTeX users please use
%\begin{thebibliography}{}
%
% and use \bibitem to create references. Consult the Instructions
% for authors for reference list style.
%
%\bibitem{RefJ}
% Format for Journal Reference
%Author, Article title, Journal, Volume, page numbers (year)
% Format for books
%\bibitem{RefB}
%Author, Book title, page numbers. Publisher, place (year)
%% etc
%\end{thebibliography}

\end{document}